%% file: root.tex
\documentclass[lettersize,journal]{IEEEtran}
\usepackage{amsmath,amsfonts}
\usepackage{algorithmic}
\usepackage{algorithm}
\usepackage{array}
\usepackage[caption=false,font=normalsize,labelfont=sf,textfont=sf]{subfig}
\usepackage{textcomp}
\usepackage{stfloats}
\usepackage{url}
\usepackage{verbatim}
\usepackage{graphicx}
\usepackage{cite}
\usepackage{gensymb}
\usepackage{setspace}
\usepackage{amsfonts} 
\usepackage{graphicx}
\usepackage{booktabs}
\usepackage{enumitem}
\usepackage{lipsum}

\begin{document}

\title{Motion Manifold Flow Primitives for Task-Conditioned \\ Trajectory Generation under Complex Task-Motion Dependencies}

\author{
 {Yonghyeon Lee$^{1}$, Byeongho Lee$^{2}$, Seungyeon Kim$^{2}$, and Frank C. Park$^{2}$}
 \thanks{$^{1}$Y. Lee is with Center for AI and Natural Sciences (CAINS), Korea Institute for Advanced Study (KIAS), Seoul 02455, South Korea.  (e-mail: ylee@kias.re.kr). $^{2}$B. Lee, S. Kim, and F. C. Park are with Robotics Laboratory, Seoul National university, Seoul 08826, South Korea.}
 \thanks{This work has been submitted to the IEEE for possible publication. Copyright may be transferred without notice, after which this version may no longer be accessible.}
 }

\markboth{Journal of \LaTeX\ Class Files,~Vol.~14, No.~8, August~2021}%
{Shell \MakeLowercase{\textit{et al.}}: A Sample Article Using IEEEtran.cls for IEEE Journals}


\maketitle

\begin{abstract}
Effective movement primitives should be capable of encoding and generating a rich repertoire of trajectories -- typically collected from human demonstrations -- conditioned on task-defining parameters such as vision or language inputs. While recent methods based on the motion manifold hypothesis, which assumes that a set of trajectories lies on a lower-dimensional nonlinear subspace, address challenges such as limited dataset size and the high dimensionality of trajectory data, they often struggle to capture complex task-motion dependencies, i.e., when motion distributions shift drastically with task variations. To address this, we introduce Motion Manifold Flow Primitives (MMFP), a framework that decouples the training of the motion manifold from task-conditioned distributions. Specifically, we employ flow matching models, state-of-the-art conditional deep generative models, to learn task-conditioned distributions in the latent coordinate space of the learned motion manifold. Experiments are conducted on language-guided trajectory generation tasks, where many-to-many text-motion correspondences introduce complex task-motion dependencies, highlighting MMFP's superiority over existing methods.
\end{abstract}

\begin{IEEEkeywords}
Movement primitives, manifold learning, flow matching models, task-conditioned motion generation
\end{IEEEkeywords}

\input{tabs/1_intro}
\input{tabs/2_related_works}
\input{tabs/3_prelim}
\input{tabs/4_mmfp}
\input{tabs/5_exp}
\input{tabs/6_con}

\bibliographystyle{IEEEtran}
\bibliography{ref}

\vfill

\end{document}

%% file: tabs/1_intro.tex
\section{Introduction}
{\it Movement primitives}, mathematical models that encode trajectories {\it offline}, enable rapid {\it online} trajectory generation~\cite{park2008movement, paraschos2013probabilistic, khansari2011learning, noseworthy2020task, lee2023equivariant, lee2024mmp++}, where data are typically collected from human demonstrations. They avoid the need for time-consuming online motion planning that reduces the system's reactivity, allowing fast adaptation to changes in environments. 

In this paper, we adopt the view that effective primitive models should satisfy two key properties. First, they should encode a diverse set of trajectories for a given task to enhance adaptability, ensuring that if some trajectories become unexecutable, such as being blocked by unforeseen obstacles, feasible alternatives remain available. Second, they should be capable of generating motions conditioned on task-defining parameters, such as vision data describing the environment or language commands reflecting user intent, ensuring applicability across a wide range of tasks.

Learning these primitives poses two main challenges.
First, obtaining a sufficiently large dataset of human demonstrations is challenging, often leading to limited data availability. This difficulty is amplified by the need to collect data for each specific task parameter, further increasing data demands.
Second, trajectory data is inherently high-dimensional, particularly for long-horizon motions. When expressed as a sequence of configurations at discrete time steps, its dimensionality increases linearly with the number of time steps.

Leveraging a low-dimensional manifold structure can be an effective solution. Assuming the set of demonstration trajectories lies on a lower-dimensional manifold embedded in the trajectory space (the manifold hypothesis), recent works have focused on identifying this manifold to reduce data dimensionality and address both challenges~\cite{noseworthy2020task, lee2023equivariant}. Task-conditioned variational autoencoders (TCVAE)~\cite{noseworthy2020task} and motion manifold primitives (MMP)~\cite{lee2023equivariant} employ conditional autoencoder architectures, demonstrating promising results for task-conditioned, diverse, high-dimensional trajectory generation.

\begin{figure}[!t]
    \centering
    \includegraphics[width=\linewidth]{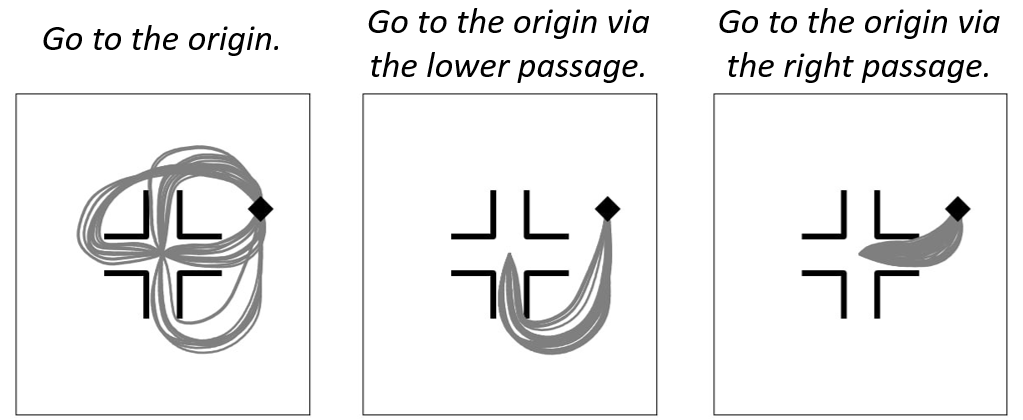}
        \caption{An illustrative example of a language-guided navigation scenario with complex task-motion dependencies: The motion distribution shifts dramatically -- for instance, in the number of modalities -- when the task parameter (in this case, a language command) changes.}
    \label{fig:example}
    \vspace{-10pt}
\end{figure}

However, existing manifold-based models fail to capture the {\it complex} task dependencies of motion distributions, where ``complex dependencies'' refer to cases in which the motion distribution undergoes dramatic shifts (e.g., changes in the number of modalities or the volume of support) as the task parameter varies.
For example, consider the navigation scenario shown in Fig.~\ref{fig:example}, where the task parameter is specified by a language command. If the command is ``Go to the origin'', the robot can select any of the four available passages. However, if the command changes to ``Go to the origin via the lower passage'', the robot must exclusively take the lower passage. This change drastically alters the motion distribution, reducing the number of modalities. As we demonstrate later in our experiments, existing models fail in these scenarios. The fundamental reason for these failures is their reliance on conditional autoencoder architectures with a shared latent prior distribution, as we detail in Section~\ref{sec:limit}.

In this paper, we propose Motion Manifold Flow Primitives (MMFP), a framework that decouples the training of the motion manifold and task-conditioned distributions, effectively capturing complex task-motion dependencies. Our approach leverages flow matching models~\cite{lipman2022flow,chen2023riemannian}, state-of-the-art conditional deep generative models capable of effectively capturing complex conditional distributions, in the latent coordinate space of the learned motion manifold. As a result, MMFP simultaneously addresses the challenges of limited datasets, high dimensionality, and complex task-motion dependencies, and significantly outperforms existing manifold-based models.

It is instructive to highlight the differences between recent learning-from-demonstration methods based on diffusion and flow matching models -- such as diffusion policies~\cite{chi2023diffusion,ze20243d} and flow matching policies~\cite{braun2024riemannian,chisari2024learning} -- and our proposed MMFP. The key distinction lies in whether the low-dimensional manifold structure is explicitly utilized. These existing methods focus on generating relatively low-dimensional \emph{local} trajectories for policy learning, whereas our aim is to produce \emph{global}, high-dimensional trajectories for motion planning, requiring dimensionality reduction. As we show in our experiments, simply training diffusion or flow matching models without manifold learning fails to generate valid global trajectories.

We provide case studies on language-guided trajectory generation, where each demonstration trajectory is paired with multiple hierarchical text descriptions.  
This setup introduces many-to-many mappings between text and motion, leading to complex text dependencies in the motion distributions.  
Two text-based trajectory generation tasks are considered: (i) SE(3) trajectory generation for a bottle in a pouring task and (ii) 7-DoF robot arm trajectory generation for a waving task. Notably, with only 10 and 30 demonstration trajectories for each task, respectively, MMFP demonstrates excellent performance.  
In contrast, diffusion and flow-based models trained directly in the high-dimensional trajectory space~\cite{ho2020denoising, lipman2022flow, chen2023riemannian} and existing motion manifold-based methods~\cite{noseworthy2020task, lee2023equivariant} fail to generate successful trajectories.

%% file: tabs/2_related_works.tex
\section{Related Works}
\label{sec:relatedworks}

\subsection{Movement primitives}
There has been extensive research on movement primitives, including dynamical systems-based approaches such as dynamic movement primitives (DMP)~\cite{saveriano2021dynamic, ijspeert2013dynamical, ijspeert2001trajectory, ijspeert2002learning2, schaal2007dynamics} and stable dynamical systems~\cite{khansari2011learning, neumann2013neural, khansari2014learning}, often formulated to guarantee the stability of the resulting closed-loop system. Additionally, methods that capture diverse motions using probabilistic formulations or manifold-based models have been explored~\cite{calinon2016tutorial, paraschos2013probabilistic, daab2024incremental, noseworthy2020task, lee2023equivariant, lee2024mmp++, beik2021learning}. 

Of particular relevance to our work are \emph{manifold-based models} and \emph{task-conditioned movement primitives}. Recent studies have explored autoencoder-based manifold learning algorithms for learning movement primitives~\cite{noseworthy2020task, beik2021learning, lee2023equivariant, lee2024mmp++}. \cite{beik2021learning} focuses on learning a manifold of {\it configurations}, followed by solving geodesic equations for motion planning. In contrast, \cite{noseworthy2020task, lee2023equivariant, lee2024mmp++} learn a manifold of {\it trajectories}, enabling direct motion sampling. Our approach aligns with the latter, as it also focuses on learning a manifold of trajectories.

Task-Parametrized Gaussian Mixture Model (TP-GMM) can produce new movements for unseen task parameters; however, it primarily addresses parameters defined as coordinate frames, making it less suitable for more general types of inputs such as vision or language~\cite{calinon2016tutorial}.  
Among the manifold-based models mentioned above, \cite{noseworthy2020task, lee2023equivariant} -- except \cite{lee2024mmp++} which does not address task-conditioned motion generation -- adopt conditional autoencoder architectures capable of handling a broader range of task parameters while encoding and generating diverse trajectories, closely aligning with our research goals. However, these methods fall short of capturing the {\it complex} task-motion dependencies within the motion distribution, a limitation we specifically address in this paper.

\subsection{Diffusion and flow matching for imitation learning}
Diffusion and flow matching models -- which fall into the category of probability flow-based models -- have recently gained significant attention as powerful deep generative models that enable efficient and stable training, demonstrating impressive performance on various conditional generation tasks~\cite{song2020denoising, song2020score, ho2022classifier, lipman2022flow, chen2023riemannian, tong2023conditional, lipman2024flow}. 
In robotics, particularly in the context of imitation learning, the use of diffusion and flow matching models for training stochastic {\it policies}, i.e., observation-conditioned action generation models, has recently gained considerable attention. Notable examples include diffusion policy~\cite{chi2023diffusion}, 3D diffusion policy~\cite{ze20243d}, and flow matching policy~\cite{chisari2024learning}, each handling different types of observations such as 2D images and 3D point clouds.

The key difference between recent diffusion or flow matching policies and our method lies in the research objective. While these approaches focus on learning a {\it local policy} that generates low-dimensional {\it local} actions -- both temporally and spatially -- often limited to a few steps of desired configurations, our goal is to develop a {\it global motion planner} capable of producing high-dimensional {\it global} trajectories. This increased dimensionality makes a naive application of diffusion or flow matching methods prone to failure, necessitating motion manifold learning as proposed in this paper. 

Moreover, while local policies can generate actions with high-frequency feedback, the produced actions often lack temporal consistency with previous ones. When actions are defined at the trajectory level, the final state of one action may not smoothly connect with the initial state of the next, leading to non-smooth behavior. For tasks where frequent trajectory re-generation is unnecessary, a global planner is generally better suited for producing long-horizon, smooth motions.

%% file: tabs/3_prelim.tex
\section{Preliminaries}
\label{sec:prelim}
This section introduces autoencoder-based manifold learning and flow matching models.

\subsection{Autoencoder-based manifold learning}
\label{sec:prelim-part1}
In this section, we briefly introduce an {\it autoencoder} and its manifold learning perspective~\cite{arvanitidis2017latent,lee2021neighborhood,lee2022statistical,yonghyeon2022regularized,janggeometrically,lee2023geometric,lee2023explicit,lim2024graph}. Consider a high-dimensional data space ${\cal X}$ and a set of data points ${\cal D}=\{x^i \in {\cal X}\}_{i=1}^{N}$, and assume the data points $\{x^i\}$ lie approximately on some lower-dimensional manifold ${\cal M}$ in ${\cal X}$ (the manifold hypothesis). 
Suppose ${\cal M}$ is an $m$-dimensional manifold and let ${\cal Z}$ be a latent space $\mathbb{R}^{m}$.

An autoencoder consists of an encoder, a mapping $g:{\cal X} \to {\cal Z}$, and a decoder, a mapping $f:{\cal Z} \to {\cal X}$. 
These are often approximated with deep neural networks and trained to minimize the reconstruction loss 
\begin{equation}
    \frac{1}{N}\sum_{i=1}^{N} d^2(f\circ g(x^i), x^i)
\end{equation}
given a distance metric $d(\cdot, \cdot)$ on ${\cal X}$. 
We note that, given a sufficiently low reconstruction error, all the data points $\{x^i\}$ should lie on the image of the decoder $f$. 
Under some mild conditions -- that are (i) $m$ is lower than the dimension of the data space ${\cal X}$ and (ii) $f$ is smooth and its Jacobian $\frac{\partial f}{\partial z}(z) \in \mathbb{R}^{dim({\cal X}) \times m}$ is full rank everywhere --, the image of $f$ is an $m$-dimensional differentiable manifold embedded in ${\cal X}$. 
In other words, the decoder produces a lower-dimensional manifold where the data points approximately lie.   

\subsection{Flow matching models} 
\label{sec:prelim-part2}
In this section, we give a brief overview of \textit{flow matching models}~\cite{lipman2022flow,lipman2024flow}, a state-of-the-art class of deep generative models.
Let $\{x^i \in {\cal X}\}_{i=1}^{N}$ be a set of data points sampled from the underlying probability density $q(x)$. 
Consider a non-autonomous vector field $v:[0,1] \times {\cal X} \to T{\cal X}$ that leads to a flow $\phi:[0,1] \times {\cal X} \to {\cal X}$ via the following ordinary differential equation (ODE): 
\begin{equation}
    \frac{d}{ds}\phi_s(x) = v_s(\phi_s(x)),     
\end{equation}
where $\phi_0(x) = x$.
Given a prior density at $s=0$ denoted by $p_0$, the flow $\phi_s$ leads to a probability density path for $s \in [0,1]$~\cite{lipman2022flow}: 
\begin{equation}
    p_s(x) = p_0(\phi_s^{-1}(x)) \det\big(\frac{\partial \phi_s^{-1}}{\partial x}(x)\big).
\end{equation}
Our objective is to learn a neural network model of $v_s(x)$ so that the flow of $v_s(x)$ transforms a simple prior density $p_0$ (e.g., Gaussian) to the target data distribution $p_1 \approx q$. Then we can sample new data points by solving the ODE, $x'=v_s(x)$, from $s=0$ to $s=1$ with initial points sampled from $p_0$. 

A key innovation in~\cite{lipman2022flow,lipman2024flow} is the \emph{flow matching loss}, which enables training \(v_s(x)\) efficiently \emph{without} simulating the ODE:
\begin{equation}
\label{eq:fmloss}
\mathbb{E}_{x_0 \sim p_0(x), \; x_1 \sim q(x), \; s \sim \mathcal{U}[0,1]}
\Bigl[
    \| v_s(x_s) - (x_1 - x_0) \|^2
\Bigr],
\end{equation}
where $x_s = (1-s)x_0 + s x_1$.
Here, \(\mathcal{U}[0,1]\) is the uniform distribution over \([0,1]\). Minimizing (\ref{eq:fmloss}) gives a vector field \(v_s(x)\) that transforms \(p_0\) into a final density \(p_1\), closely approximating the data distribution \(q\).

We are particularly interested in a \emph{conditional} density function \(p(x | c)\), where \(c\) is a conditioning variable. To address this, a neural network vector field \(v_s(x, c)\) takes both \(x\) and \(c\) as inputs, and is trained using a similar objective:
\begin{equation}
\label{eq:cfmloss}
\mathbb{E}_{x_0 \sim p(x_0), \; (x_1,c) \sim q(x,c), \; s \sim \mathcal{U}[0,1]}
\Bigl[
    \| v_s(x_s, c) - (x_1 - x_0) \|^2
\Bigr],
\end{equation}
where $x_s = (1-s)x_0 + s x_1$, and \(q(x,c)\) is the underlying joint distribution of data points and their conditioning variables. By incorporating \(c\) into the vector field, the flow can adapt to various context inputs, enabling conditional sample generation.

%% file: tabs/4_mmfp.tex
\section{Motion Manifold Flow Primitives} 
We begin this section by introducing the notations and assumptions used throughout the paper. Let \({\cal Q}\) be a configuration space, and denote a sequence of configurations by \(x = (q_1, \ldots, q_T)\), referred to as a trajectory, where \(q_i \in {\cal Q}\) for all \(i\) and \(T\) is a fixed positive integer representing the length of trajectory. The trajectory space is then denoted by \({\cal X} = {\cal Q}^T\), which is typically very high-dimensional.
We denote the task parameter by \(c \in {\cal C}\) and assume access to a trajectory-task pair dataset \({\cal D} = \{(x^i, c^i)\}_{i=1}^{N}\), where many-to-many correspondences exist between \(x\) and \(c\) (e.g., multiple values of \(c\) may correspond to the same \(x\) and vice versa).

In the subsequent sections, we first explain the limitations of existing task-conditioned motion manifold primitives, introduce our Motion Manifold Flow Primitives (MMFP), and describe its specific application when the task parameter is a free-form text input.

\subsection{Limitations of existing motion manifold primitives}
\label{sec:limit}
Existing motion manifold-based models, TCVAE~\cite{noseworthy2020task} and MMP~\cite{lee2023equivariant}, adopt conditional autoencoder architectures for task-conditioned trajectory generation. Both assume a lower-dimensional latent space \({\cal Z}\) and train an encoder \(g : {\cal X} \to {\cal Z}\) and a conditional decoder \(f : {\cal Z} \times {\cal C} \to {\cal X}\) (or their stochastic variants) that map a trajectory \(x\) to a latent representation \(z\) and generate a trajectory given \((z, c)\). The latent prior \(p_{\rm prior}(z)\) is typically modeled as a Gaussian or a Gaussian mixture model. Motion sampling involves drawing \(z\) from the prior and mapping it through the decoder \(f\).

We note that the latent prior is {\it shared} across all task parameters, requiring the conditional decoder to sufficiently distort the same prior into different motion distributions based on the task parameter. However, when motion distributions shift dramatically (complex task-motion dependencies), the decoder often struggles to capture the complex nonlinear transformations required. In contrast, probability flow-based models continuously transform the prior via an ODE, where smooth, incremental changes accumulate into significant variations, making them better suited for modeling such complex dependencies.

\subsection{Decoupling manifold learning and conditional densities}
Our Motion Manifold Flow Primitives (MMFP) framework consists of two modules: (i) motion manifold and (ii) latent flow.
The motion manifold model consists of two neural networks, an encoder $g:{\cal X} \to {\cal Z}$ and a decoder $f:{\cal Z} \to {\cal X}$, which are trained as follows:
\begin{equation}
    \min_{f,g} \frac{1}{N}\sum_{i=1}^{N} d^2(x^i, f(g(x^i))) + \eta \| g(x^i)\|^2 + \delta {\cal E}(f,g),
\end{equation}
where $\eta,\delta$ are some positive scalars. The second term penalizes the norm of latent values to prevent them from diverging excessively far from the origin. The third term is added to ensure their smoothness, which is defined as follows: ${\cal E}(f,g):= \mathbb{E}_{z} \big[ \sum_{t=1}^{T-1} \| f^{t+1}(z) - f^t(z)\|^2 \big]$, where $f(z) = (f^1(z), \ldots, f^T(z)) \in {\cal Q}^T$.
The latent value \( z \) in this expectation is sampled with augmentation as \( z = \alpha g(x^i) + (1-\alpha) g(x^j) \), where \( \alpha \sim {\cal U}[-0.4, 1.4] \) and \( x^i, x^j \) are sampled from the dataset, to extend the regularization effect to regions where data is not available, adopting~\cite{yonghyeon2022regularized}.

Given a pre-trained encoder \(g\), generate a set of latent-task pairs \(\{(z^i, c^i)\}_{i=1}^N\) where \(z^i = g(x^i)\). A neural network vector field \(v_s(z, c)\) is then trained using the flow matching loss in (\ref{eq:cfmloss}). Finally, given a task parameter \(c\), motion sampling is performed by integrating the ODE \(dz/ds = v_s(z, c)\) from \(s=0\) to \(s=1\) to obtain latent samples \(z_1\) (where the index denotes the final time step \(s=1\)), followed by decoding them using \(f : z_1 \mapsto x\).

A reasonable question may arise: Why not use diffusion models~\cite{song2020score}, which are modeled with stochastic differential equations (SDEs), instead of flow matching models based on ordinary differential equations (ODEs), in the latent space? One answer is that while both approaches can capture complex conditional dependencies, flow matching models enable faster generation~\cite{lipman2022flow}.
Moreover, given a limited dataset size, flow matching tends to achieve better interpolation and generalization compared to diffusion models, as optimal transport paths used in flow matching are smoother than the diffusion paths~\cite{lipman2022flow}. Our empirical studies further support this, showing that latent flow models outperform latent diffusion models; see Section~\ref{sec:toy_exp}.

\subsection{Language as a task parameter}
\begin{figure*}
    \centering
    \includegraphics[width=\linewidth]{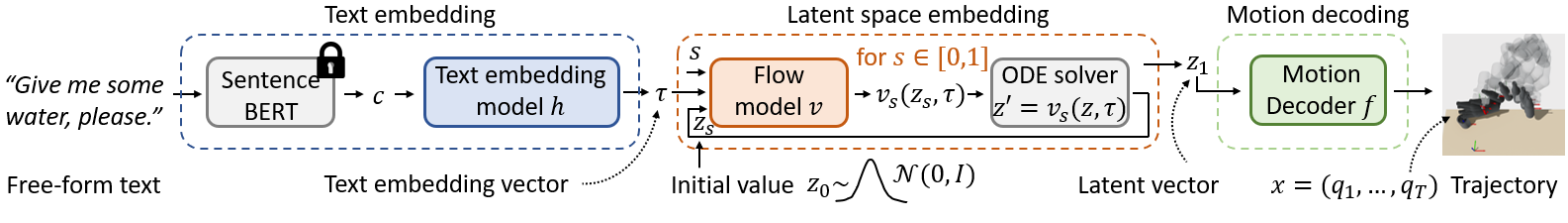}
    \caption{The procedure of motion generation in MMFP: (i) the Sentence-BERT encodes a free-form text into a vector $c$, (ii) the text embedding model $h$ maps $c$ to a text embedding vector $\tau$, (iii) we solve the ODE $z'=v_s(z,\tau)$ from $s=0$ to $s=1$ with an initial value $z_0$ sampled from Gaussian ${\cal N}(z|0,I)$ and obtain $z_1 \in {\cal Z}$, and (iv) the motion decoder $f$ maps $z_1$ to a trajectory $x=(q_1,\ldots,q_T)$.}
    \label{fig:method}
\end{figure*}

We use a pre-trained text encoder, Sentence-BERT~\cite{reimers-2019-sentence-bert}, without fine-tuning, to encode free-form texts into 768-dimensional vectors. These vectors serve as task parameters, denoted by \(c\). However, we empirically observed that directly inputting the 768-dimensional task parameter into the velocity model \(v_s(z,c)\) does not produce satisfactory results. To address this limitation, we introduce a text embedding model, a neural network \(h\), which encodes \(c\) into a lower-dimensional vector \(\tau \in \mathcal{T}\).

We train the text embedding model $h: c \mapsto \tau$ and velocity field $v: (s,z,\tau) \mapsto dz/ds$ simultaneously with the following {\it regularized} flow matching loss:

{\small
\begin{equation}
   \label{eq:lfl}
   \sum_{i,k} \mathbb{E}_{s,z_0} \big[\|v_s(z_s^i, h(c^i)) - (z^i - z_0)\|^2\big] + \gamma \|h(c^i) - h(\tilde{c}^{ik})\|^2 ,
\end{equation}
}
where $z_s^i = (1-s)z_0 + s z^i$, and $s, z_0$ are sampled from $\mathcal{U}[0,1]$ and a Gaussian prior, respectively. The second term, scaled by a positive weight \(\gamma\), encourages robustness against diverse text variations. Here, \(\tilde{c}^{ik}\) for $k=1,\ldots,K$ are Sentence-BERT encoding vectors with meanings similar to those of \(c^i\), generated using the Large Language Model \textit{ChatGPT}. 
We refer to Fig.~\ref{fig:method} for an overview of the motion sampling procedure using MMFP given a free-form text input.

%% file: tabs/5_exp.tex
\section{Experiments}
\label{sec:experiments}
In this section, we evaluate our method, MMFP, primarily compared to (i) Denoising Diffusion Probabilistic Models (DDPM)~\cite{ho2020denoising}, (ii) Flow Matching (FM)~\cite{lipman2022flow}, (iii) Task-Conditional Variational Autoencoder (TCVAE) with a Gaussian prior~\cite{noseworthy2020task}, and (iv) Motion Manifold Primitives with a Gaussian mixture prior (MMP)~\cite{lee2023equivariant}.  
DDPM and FM are directly trained in the trajectory space \(\mathcal{X}\). When \(\mathcal{X} = \mathrm{SE}(3)^T\), a Riemannian manifold, we train a DDPM in local coordinates (using exponential coordinates for the $\mathrm{SO}(3)$ component) and use Riemannian Flow Matching (RFM)~\cite{chen2023riemannian}.  
MMFP trained without the robustness regularization term (the second term in (\ref{eq:lfl})) will be denoted as MMFP w/o reg.

For MMFP, we choose the latent space dimension $m=3$ for ${\cal Z} = \mathbb{R}^{m}$ and text embedding dimension $p=3$ for ${\cal T}=\mathbb{R}^{p}$. For TCVAE and MMP, with thorough tuning, we set the latent space dimension to be $3$ or $4$ and text embedding dimension to be $64$.
We compare models for 6-DoF SE(3) pouring trajectory generation and 7-DoF robot arm waving motion generation tasks. 
Throughout, we use fully-connected neural networks with ELU activation functions.

\textbf{Evaluation metrics:} {The primary objective is to generate correct motions corresponding to the given text inputs. To evaluate whether the model generates accurate motions that perform the task described in the given text, we train a trajectory classifier and use it to report the motion accuracy; the higher, the better. 
Additionally, the model should be able to generate not just a single trajectory but diverse trajectories, imitated from the given demonstration dataset, depending on the text input. For example, see Fig.~\ref{fig:diverse_pouring_motion_corl}. 
We use the Maximum Mean Discrepancy (MMD)~\cite{gretton2012kernel} -- which measures the distance between two probability distributions, computed from their samples -- to measure the similarity between the set of generated trajectories given a text input and the set of demonstration trajectories annotated with that text; the lower, the better. 

Texts are annotated to each trajectory at multiple different levels (e.g., see Fig.~\ref{fig:pouring_dataset_and_results_corl} {\it Left} and Fig.~\ref{fig:dataset_waving} {\it Left}), with higher levels specifying more detailed requirements. Multiple trajectories can correspond to the same text (e.g., all trajectories being assigned the same level 1 text), resulting in many-to-many text-motion correspondences.
The MMD metrics are measured separately for each level text input, then averaged across text descriptions at the same level. 
Evaluation metrics are computed using both seen and unseen texts in training. First, accuracy and MMD metrics are measured using the seen texts. Then, to evaluate the robustness to text variations, we also report robust MMD metrics, which are computed with unseen text inputs generated by ChatGPT.

We begin this section by comparing latent diffusion and flow matching models on text-based 2D motion generation tasks. We then present experiments involving higher-dimensional configuration spaces, including text-based SE(3) pouring motion generation and 7-DoF waving motion generation tasks.

\subsection{Latent diffusion vs latent flow matching}
\label{sec:toy_exp}
In this section, using the simple 2D trajectory generation task shown in Fig.~\ref{fig:toy_data}, we compare our motion manifold flow primitives trained with latent flow matching models to those trained with latent diffusion models~\cite{song2020score}, denoted as MMP + Diffusion, while using the same motion manifold (an autoencoder). 

The dataset consists of 20 demonstration trajectories, all starting from the same point and ending at a common goal (the origin). Each trajectory has a length of $T=201$. For each trajectory, two text annotations are provided (see Fig.~\ref{fig:toy_data}). 

Diffusion models involve several design choices~\cite{song2020denoising}: the diffusion paths and noise schedules. In this experiment, latent diffusion models are trained using various Gaussian probability paths, including the variance-exploding path, denoted as Diffusion-VE, and two variance-preserving paths with different noise schedules, denoted as Diffusion-VP-1 and Diffusion-VP-2.

As shown in Table~\ref{tab:main_toy_extended} and Fig.~\ref{fig:appendix_dif_vs_ot}, MMFP achieves strong performance across both levels of text inputs simultaneously. In contrast, MMP combined with diffusion models yields less satisfactory results, performing well on either the level 1 or level 2 task but failing to achieve strong performance on both simultaneously.
These results can be attributed to the smoother optimal transport paths~\cite{lipman2022flow} used in flow matching compared to diffusion paths. These smoother paths simplify the underlying ground-truth vector field that the neural network vector field \(v_s(z,c)\) needs to fit, making the fitting process easier, improving accuracy, and reducing the number of function evaluations required during sampling.

\begin{figure}[!t]
    \centering
    \includegraphics[width=0.3\textwidth]{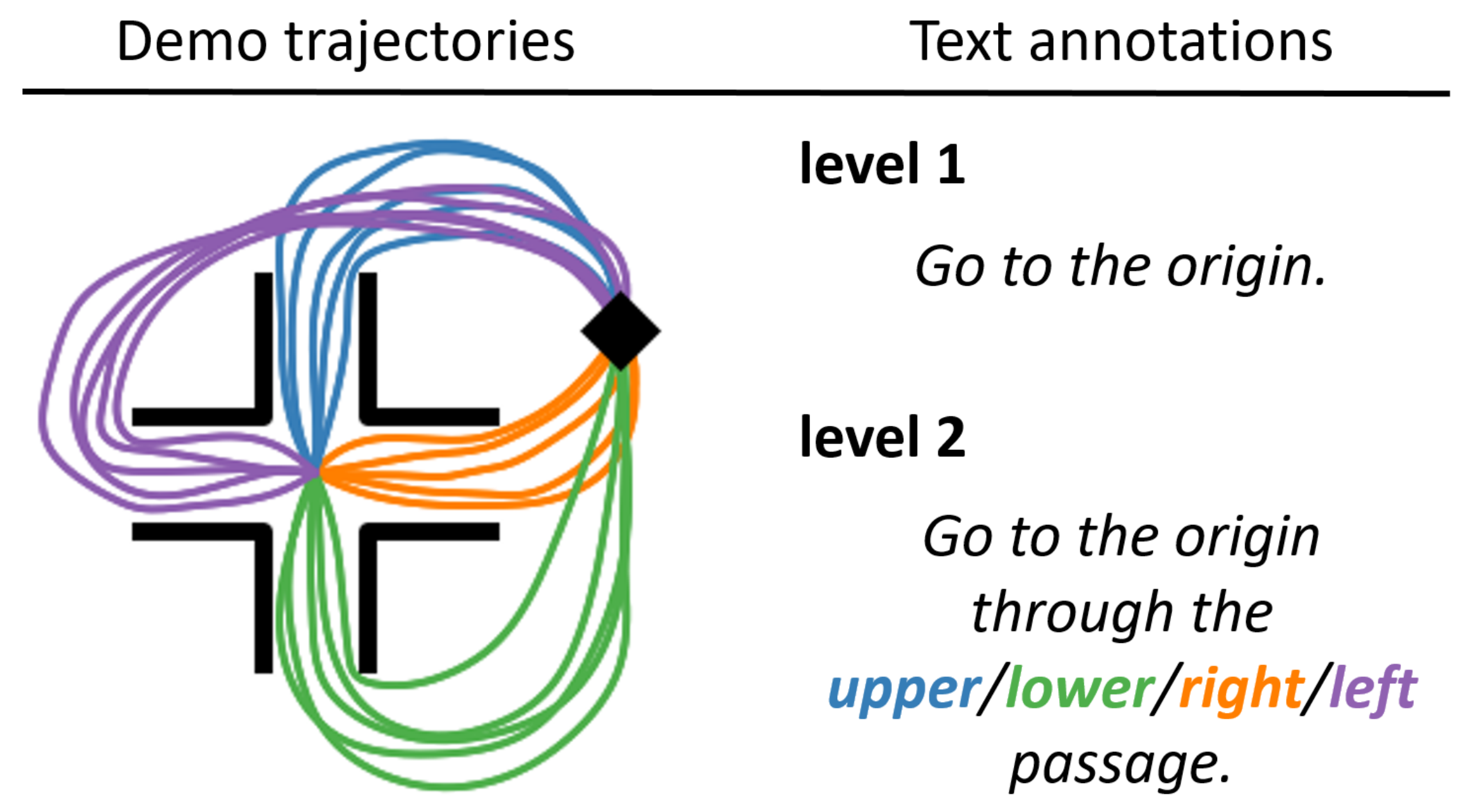}
    \caption{Demonstration trajectories with multiple text annotations. Each trajectory is assigned with two text labels, the level 1 and level 2 texts.}
    \label{fig:toy_data}
\end{figure}

\begin{table}[!t]
\centering
\scriptsize
\caption{The MMD metric and accuracy ($\%$) for 2D motion generation.}
\label{tab:main_toy_extended}
\begin{tabular}{ccccc} 
\toprule
& \multicolumn{2}{c}{MMD ($\downarrow$)}  & \multicolumn{2}{c}{Accuracy ($\uparrow$)}  \\ \cmidrule(lr){2-3} \cmidrule(lr){4-5} 
& level 1 & level 2 & \begin{tabular}[c]{@{}c@{}} path \end{tabular} & task \\ 
                                                           \midrule
\begin{tabular}[c]{@{}c@{}}MMP + Diffusion-VE \end{tabular}  & 0.075 & {\bf 0.007} & 100 & 94.6 \\
\begin{tabular}[c]{@{}c@{}}MMP + Diffusion-VP-1 \end{tabular} & 0.073 & {\bf 0.003} & 100 & 95.0   \\
\begin{tabular}[c]{@{}c@{}}MMP + Diffusion-VP-2 \end{tabular} & {\bf 0.030} & 0.055 & 94.3 & 81.0   \\
\begin{tabular}[c]{@{}c@{}}MMFP \end{tabular}  & {\bf 0.025} & {\bf 0.004} & {\bf 100} & {\bf 99.8}   \\
\bottomrule
\end{tabular}
\end{table}

\begin{figure}[!t]
    \centering
    \includegraphics[width=1\linewidth]{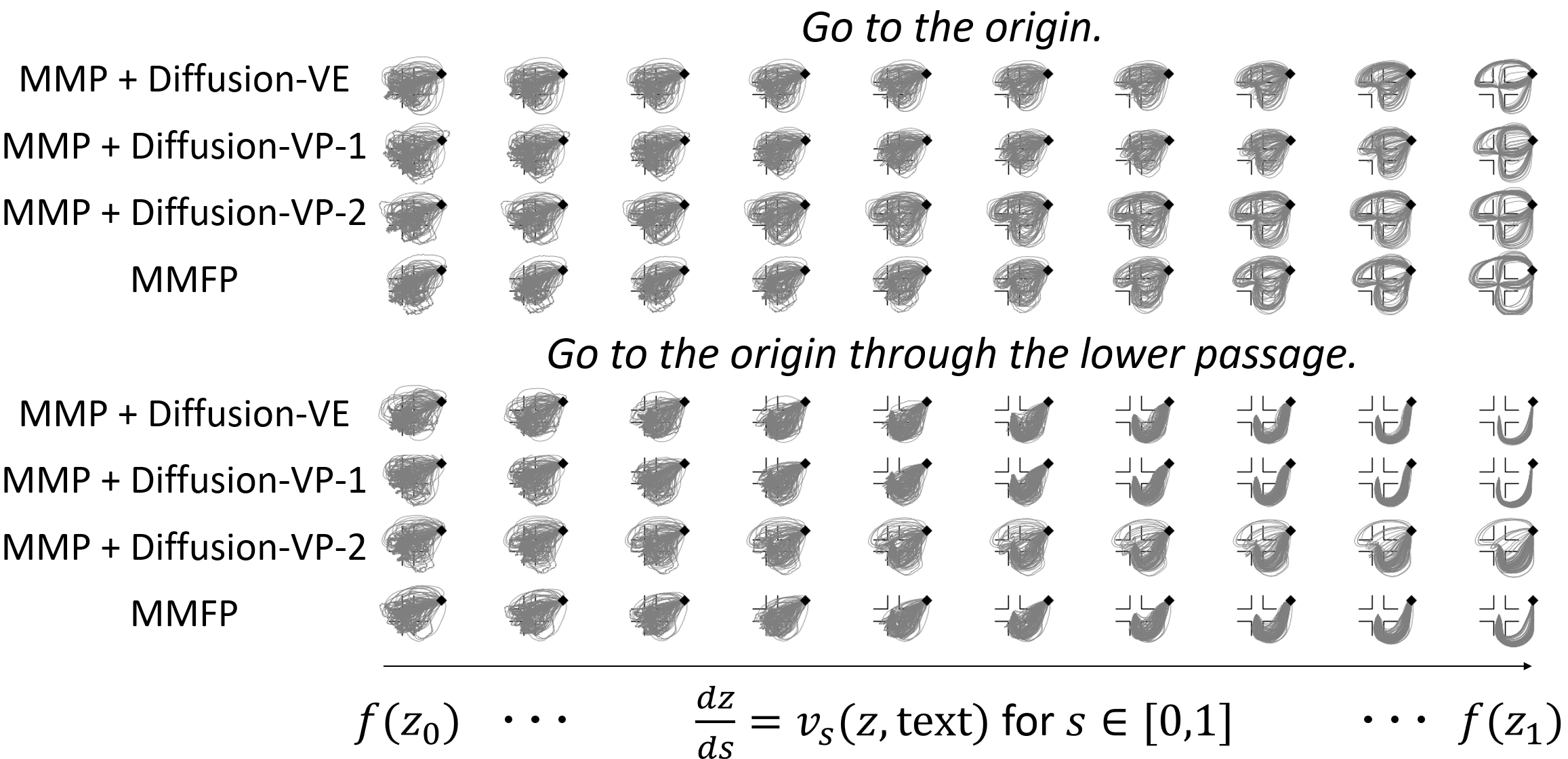}
    \caption{The evolution of generated trajectories (from left to right) follows each of the latent models, either diffusion or flow, with the trajectories on the far right representing the final output samples.}
    \label{fig:appendix_dif_vs_ot}
\end{figure}

\subsection{SE(3) pouring motion generation}

\begin{table*}[!t]
\scriptsize
\centering
\caption{The MMD and robust MMD metrics and the accuracy ($\%$) for pouring motion generation.}
\label{tab:main}
\begin{tabular}{lccccccccc} 
\toprule
& \multicolumn{3}{c}{MMD ($\downarrow$)}    & \multicolumn{3}{c}{robust MMD ($\downarrow$)} & \multicolumn{3}{c}{Accuracy ($\uparrow$)}  \\ 
\cmidrule(lr){2-4} \cmidrule(lr){5-7} \cmidrule(lr){8-10}
& level 1 & level 2 & level 3 & level 1   & level 2  & level 3  & \begin{tabular}[c]{@{}c@{}}pouring \\ style\end{tabular} & \begin{tabular}[c]{@{}c@{}}pouring \\ direction\end{tabular} & both \\ 
\midrule
DDPM~\cite{ho2020denoising} & 0.398 & 0.484 & 1.306 & 0.400 & 0.486 & 1.304 & 50.5 & 20.0 & 10.2\\
RFM~\cite{chen2023riemannian}   & 0.411 & 0.431 &	0.778 &	0.413 &	0.425 &	1.127 & 87.0 & 36.4 & 30.8 \\
TCVAE~\cite{noseworthy2020task} & 0.117 & 0.211 & 0.824 & 0.131 & 0.191 & 1.041 & 52.0 & 25.4 & 15.3 \\
MMP~\cite{lee2023equivariant}  & {\bf 0.045} & {\bf 0.115} & 0.950 & {\bf 0.055} & {\bf 0.112} & 0.970 & 47.5 & 19.6 & 9.3 \\
\begin{tabular}[c]{@{}c@{}} MMFP w/o reg (ours) \end{tabular} 
      & {\bf 0.055}  & {\bf 0.114} & {\bf 0.009} & {\bf 0.056} & {\bf 0.097} & 0.096 & {\bf 98.5} & {\bf 93.2} & {\bf 99.9} \\
\begin{tabular}[c]{@{}c@{}} MMFP (ours) \end{tabular} 
      & {\bf 0.042}  & {\bf 0.093} & {\bf 0.007} & {\bf 0.052} & {\bf 0.094} & {\bf 0.016} & {\bf 99.0} & {\bf 92.6} & {\bf 99.9} \\
\bottomrule
\end{tabular} 
\end{table*}

\begin{figure*}[!t]
    \centering
    \includegraphics[width=1\linewidth]{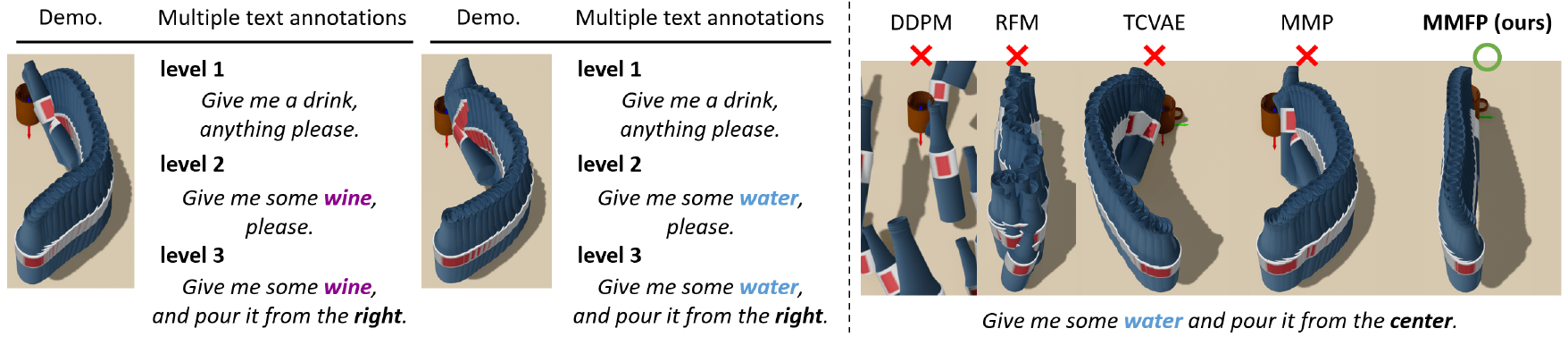}
    \caption{{\it Left}: Example demonstration trajectories, each of which is annotated with three different level texts. {\it Right}: Generated pouring trajectories by RFM, TCVAE, MMP and MMFP.}
    \label{fig:pouring_dataset_and_results_corl}
\end{figure*}

\begin{figure*}[!t]
    \centering
    \includegraphics[width=1\linewidth]{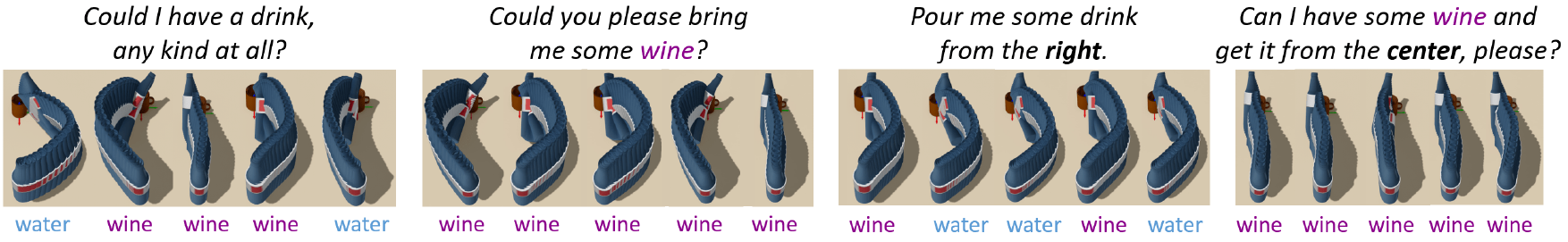}
    \caption{A variety of accurate trajectories generated by MMFP given unbiased text inputs.}
    \label{fig:diverse_pouring_motion_corl}
\end{figure*}

\begin{figure}[!t]
    \centering
    \includegraphics[width=1\linewidth]{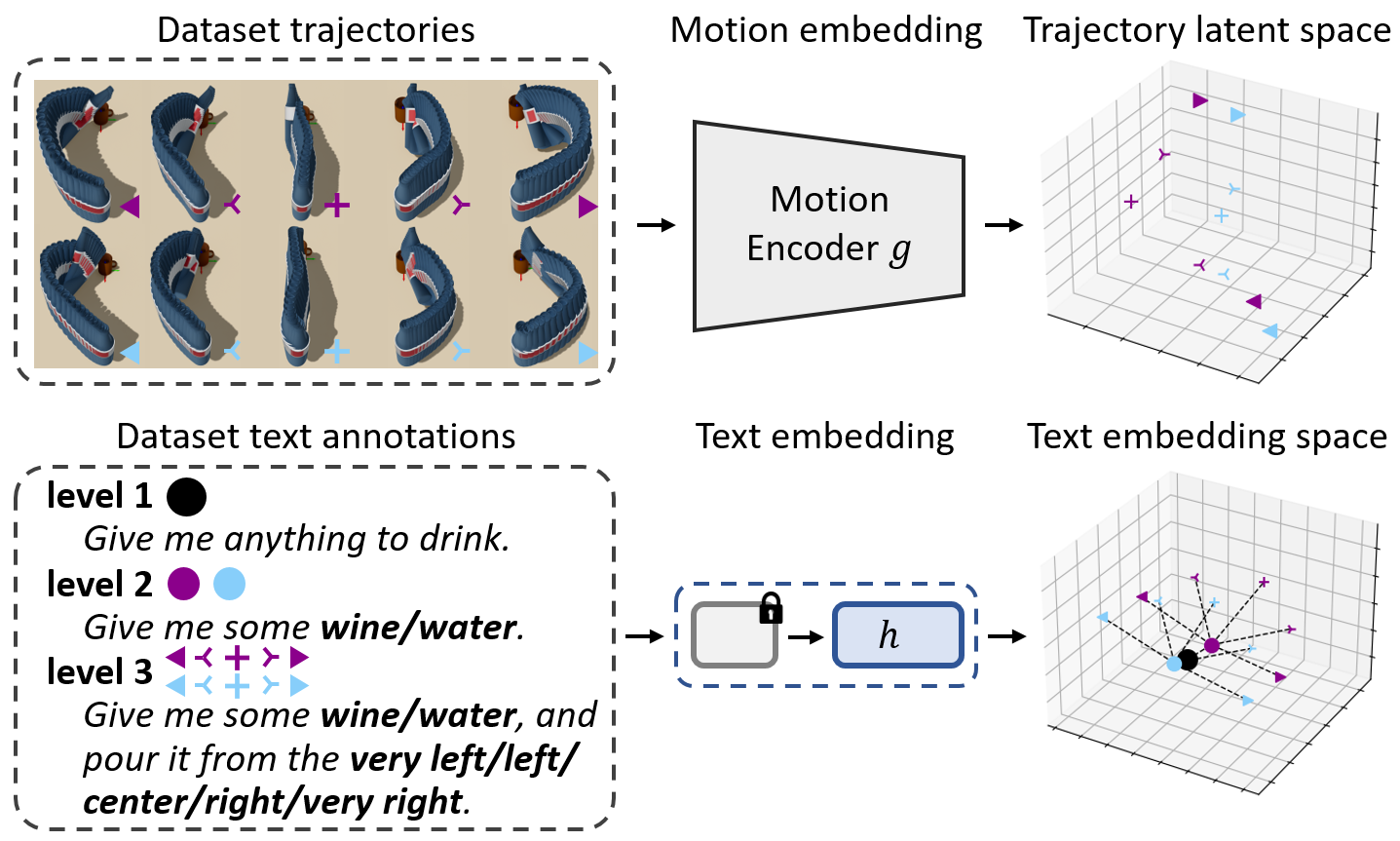}
    \caption{{\it Upper}: SE(3) pouring trajectory data encoded in the three-dimensional latent space by our motion encoder. {\it Lower}: Texts used in training encoded in the three-dimensional text embedding space by our text embedding model.}
    \label{fig:results1-label}
\end{figure}

\begin{figure}[!t]
    \centering
    \includegraphics[width=1\linewidth]{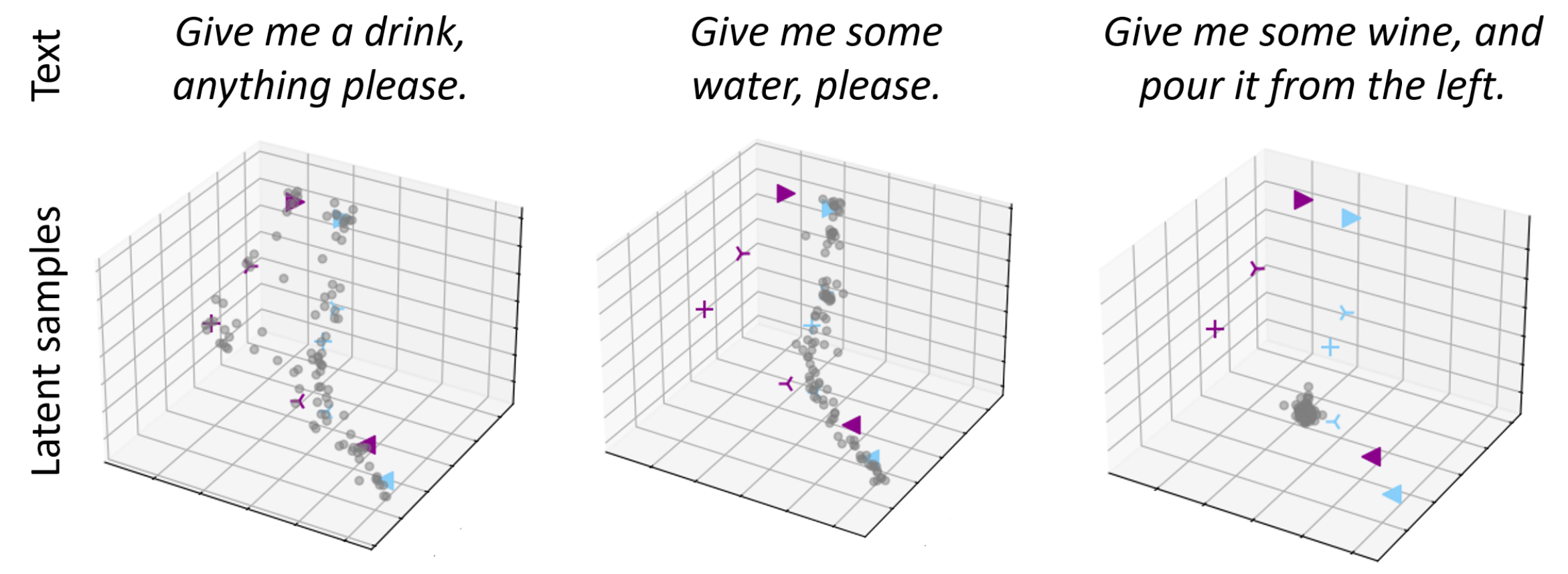}
    \caption{Sampled latent points by our trained latent flow model in MMFP. {\it Left}: Sampled latent points given the level 1 text. {\it Middle}: Sampled latent point given a level 2 text. {\it Right}: Sampled latent points given a level 3 text.}
    \label{fig:results2-label}
\end{figure}

\begin{figure}[!t]
    \centering
    \includegraphics[width=1\linewidth]{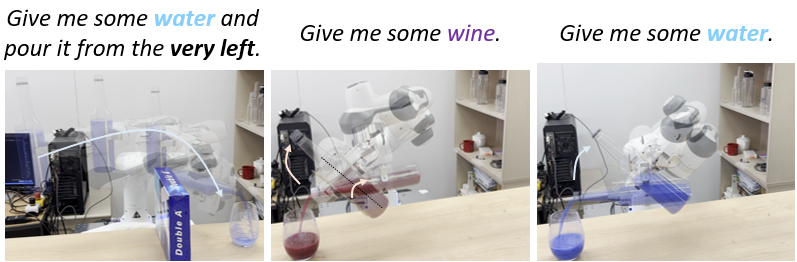}
    \caption{Real robot pouring motion generation results.}
    \label{fig:real_exp}
\end{figure}

In this section, we train text-based pouring motion generation models, where the dataset is obtained from the human demonstration videos. 
The demonstrator is instructed to pour water or wine in five different pouring directions (i.e., from the very left, left, center, right, and very right side). When pouring wine, the demonstrator is instructed to turn the wrist clockwise at the end.
From ten videos, we extract SE(3) trajectories of the bottles, and the trajectory lengths are pre-processed so that $T=480$. 
For each trajectory, we give three text annotations, as shown in Fig.~\ref{fig:pouring_dataset_and_results_corl} ({\it Left}). 
Notably, multiple trajectories can be associated with a single text label, leading to many-to-many correspondences between text and motion. For example, all 10 trajectories are labeled with the level 1 text, ``Give me a drink, anything please''.

Table~\ref{tab:main} shows MMD, robust MMD, and Accuracy; our MMFP only shows good scores in all metrics. We note that the regularization in MMFP significantly improves the level 3 robust MMD metric. 
The DDPM and RFM overall produce very poor MMD results. As shown in Fig.~\ref{fig:pouring_dataset_and_results_corl} ({\it Right}), the trajectory generated by DDPM does not even converge near the cup and that of RFM is very jerky. 
TCVAE and MMP, both adopting the motion manifold hypothesis, produce motions of reasonable quality. 
Nevertheless, they exhibit a limitation in understanding level 3 text descriptions, particularly regarding pouring directions. 
Examples illustrating this limitation can be observed in Fig.~\ref{fig:pouring_dataset_and_results_corl} ({\it Right}), where they fail in pouring from accurate directions. Lastly, Fig.~\ref{fig:diverse_pouring_motion_corl} shows a variety of accurately generated motions by MMFP given unbiased user text inputs not seen during training.

Furthermore, we present a visual analysis of our MMFP, wherein the three-dimensional nature of both the latent space \(\mathcal{Z}\) and the text embedding space \(\mathcal{T}\) facilitates straightforward visualization. 
Fig.~\ref{fig:results1-label} ({\it Upper}) shows the latent values of the trajectory data, i.e., \(z^i = g(x^i) \in \mathbb{R}^{3}\), where \(g\) is a trained encoder.  
Fig.~\ref{fig:results1-label} ({\it Lower}) displays the embedding vectors of the text inputs, i.e., \(\tau^i = h(c^i) \in \mathbb{R}^{3}\), where \(h\) is a trained text embedding module.
Semantically similar trajectories and texts are positioned closely together, exhibiting smooth transitions as their meanings shift, indicating successful model training.

Fig.~\ref{fig:results2-label} highlights the multi-modality of the text-conditioned latent distribution and, more importantly, the complex text-motion dependencies, as evidenced by the substantial shifts in the distribution with varying text inputs. In the figure, moving from left to right corresponds to scenarios where level 1 to level 3 texts are given as inputs. When a level 1 text is provided, the distribution must fit two modalities, and the volume of the distribution is the largest. In contrast, when level 2 or level 3 texts are provided, there is only one modality. Note that the support of the distribution varies significantly from level 1 to level 3, becoming much smaller at level 3.
This implies that a latent text-conditioned distribution should capture such dramatic changes, fitting a complex, non-smooth function. Existing manifold-based models relying on conditional autoencoders struggle to capture such a complex function, whereas our MMFP effectively models it.

Lastly, we conduct real robot experiments (see Fig.~\ref{fig:real_exp}). Given the generated SE(3) trajectories of the bottle, we compute joint space trajectories by solving the inverse kinematics. 
The generated SE(3) trajectory is smooth and densely discretized enough that, despite the relatively high speed of the bottle, the robot can accurately track the desired end-effector pose using computed torque control with a PD control law.

\subsection{7-DoF waving motion generation}
\begin{table*}[!t]
\centering
\scriptsize
\caption{The MMD and robust MMD metrics and the accuracy ($\%$) for waving motion generation.}
\label{tab:main_waving}
\begin{tabular}{lccccccccc} 
\toprule
& \multicolumn{3}{c}{MMD ($\downarrow$)}    & \multicolumn{3}{c}{robust MMD ($\downarrow$)} & \multicolumn{3}{c}{Accuracy ($\uparrow$)}  \\ 
\cmidrule(lr){2-4} \cmidrule(lr){5-7} \cmidrule(lr){8-10}
& level 1 & level 2 & level 3 & level 1   & level 2  & level 3  & \begin{tabular}[c]{@{}c@{}}waving \\ direction\end{tabular} & \begin{tabular}[c]{@{}c@{}}waving \\ style\end{tabular} & both \\ 
\midrule
DDPM~\cite{ho2020denoising} & 0.425 & 0.542 & 0.831 & 0.427 & 0.542 & 0.831 & 18.4 & 29.3 & 4.1 \\
FM~\cite{lipman2022flow}  & 0.211 & 0.368 & 0.646 & 0.215 & 0.384 & 0.671 & {\bf 99.6} & 88.7 & 91.1 \\
TCVAE~\cite{noseworthy2020task} & 0.269 & 0.581 & 0.833 & 0.294 & 0.606 & 0.865 & 25.6 & 57.3 & 10.7 \\
MMP~\cite{lee2023equivariant} & {\bf 0.013} & 0.369 & 0.772 & {\bf 0.016} & 0.355 & 0.750 & 19.2 & 30.7 & 6.5\\
\begin{tabular}[c]{@{}c@{}} MMFP w/o reg (ours) \end{tabular} 
      & {\bf 0.020} & {\bf 0.037} & {\bf 0.006} & {\bf 0.024} & {\bf 0.046} & 0.021 & {\bf 99.8} & {\bf 98.7} & {\bf 99.7} \\
\begin{tabular}[c]{@{}c@{}} MMFP (ours) \end{tabular} 
      & {\bf 0.016} & {\bf 0.040} & {\bf 0.004} & {\bf 0.022} & {\bf 0.040} & {\bf 0.005} & {\bf 100} & {\bf 97.7} & {\bf 100} \\
\bottomrule
\end{tabular}
\end{table*}

\begin{figure*}[!t]
    \centering
    \includegraphics[width=1\linewidth]{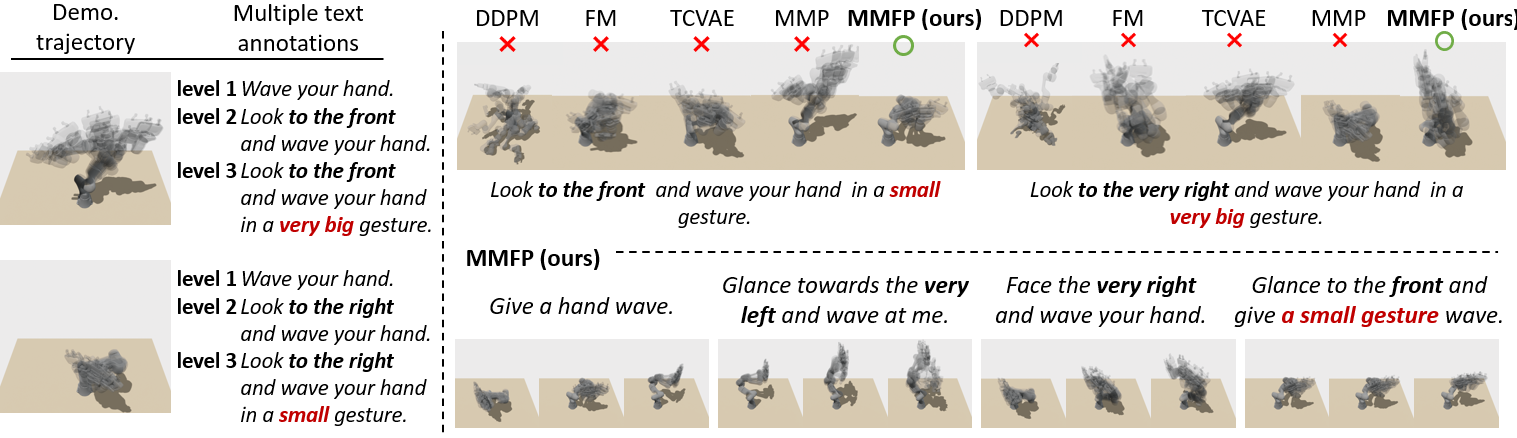}
    \caption{{\it Left}: Example 7-DoF demonstration trajectories, each of which is assigned with three different text labels. {\it Right-Upper}: Generated waving trajectories by DDPM, FM, TCVAE, MMP and MMFP. {\it Right-Lower}: A variety of accurate trajectories generated by MMFP given unbiased text inputs.}
    \label{fig:dataset_waving}
\end{figure*}

\begin{figure}[!t]
    \centering
    \includegraphics[width=0.6\linewidth]{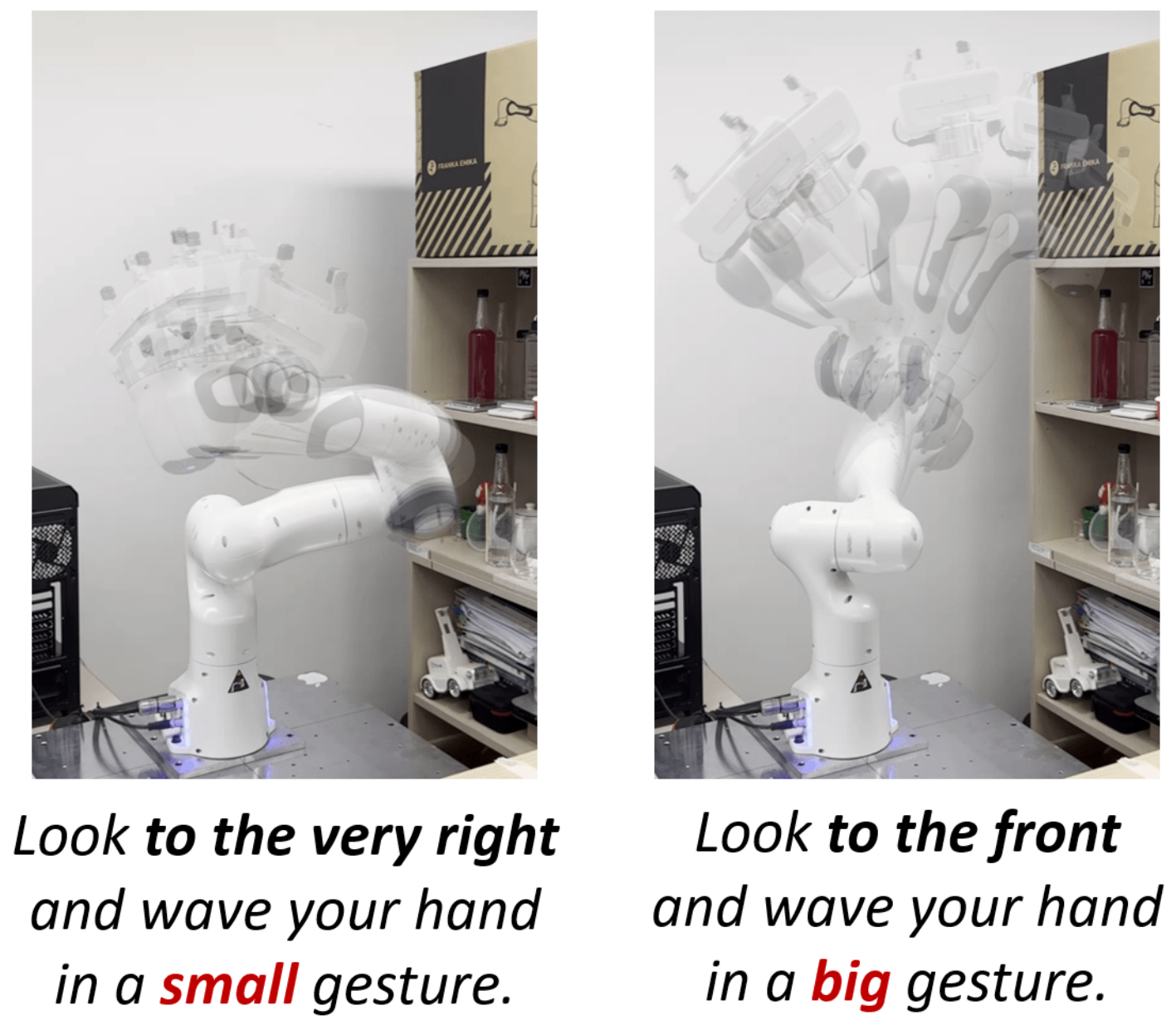}
    \caption{Real robot waving motion generation by MMFP.}
    \label{fig:real_waving_exp_corl}
\end{figure}

In this section, we train text-based 7-DoF waving motion generation models, where the dataset is obtained from human demonstrations. The demonstrator is instructed to hold and move the robot arm so that the robot arm mimics waving motions in five different viewing directions (i.e., very left, left, front, right, and very right) and in three different styles (i.e., very big, big, small). We extract 7-DoF joint space trajectories, and the trajectory lengths are pre-processed to ensure $T=720$. 
Thus, ${\cal Q} = \mathbb{R}^{7}$, and the trajectory space is denoted as ${\cal X} = \mathbb{R}^{7 \times T}$. 
We encode two motions for each setting, resulting in a total of $N=30$ trajectories. For each trajectory, we provide three text annotations, as shown in Fig.~\ref{fig:dataset_waving} ({\it Left}). 

Table~\ref{tab:main_waving} presents the MMD, robust MMD, and Accuracy metrics, where MMFP once again achieves strong performance across all measures. These results align consistently with the findings from the previous pouring experiment. See Fig.~\ref{fig:dataset_waving} for qualitative results. Additionally, examples of real robot waving motion generated using our MMFP are illustrated in Fig.~\ref{fig:real_waving_exp_corl}.

%% file: tabs/6_con.tex
\section{Conclusion}
\label{sec:conclusion}
The Motion Manifold Flow Primitive (MMFP) framework combines motion manifolds and latent flows, learning task-conditioned motion distributions with complex task-motion dependencies, all while relying on a small number of trajectory data. The performance has been validated through extensive experiments involving the language-guided generation of SE(3) and 7-DoF joint trajectories. Our results demonstrate accurate motion generation even for unseen text inputs generated by the Large Language Model, ChatGPT. This contrasts with existing manifold-based methods, such as TC-VAE and MMP, which struggle with text inputs containing detailed descriptions, and trajectory space diffusion and flow-based models, which tend to produce jerky trajectories.

We believe the MMFP framework can be further improved in several ways. The current implementation relies solely on text inputs. Extending it to incorporate visual inputs, such as depth images or point clouds representing object geometry, could make the primitives more versatile. Additionally, the framework currently requires fixed-length discrete-time trajectories, which could be addressed by adopting sequence models, such as transformers or recurrent neural networks, or continuous-time representations with parametric curve models, as proposed in~\cite{lee2024mmp++}.